\documentclass{article}

\usepackage[T1]{fontenc}
\usepackage[utf8]{inputenc}
\usepackage{microtype}
\usepackage{graphicx}
\usepackage{amsmath,amssymb}
\usepackage{booktabs}
\usepackage{multirow}
\usepackage{xspace}
\usepackage{capt-of} % Captions outside floats, using the conference caption style.
\usepackage{hyperref}

\usepackage[accepted]{mlsys2025}
\mlsystitlerunning{ChainFlow: Few-Step Latent Flow Matching for Speculative Decoding}
\mlsyskeywords{Speculative decoding, flow matching, efficient LLM inference}

\makeatletter
\renewcommand{\Notice@String}{\textsuperscript{\textit{Q}}Equal contribution.
\textsuperscript{*}Corresponding author.\\[\baselineskip] Preprint.}
\newcommand{\cfAffMark}[1]{\ifnum#1=1 \textit{K}\else\ifnum#1=2 \textit{V}\else\number#1\fi\fi}
\renewcommand{\@pa}[1]{%
  \ifcsname the@affil#1\endcsname\else
    \ifcsname @icmlsymbol#1\endcsname\else
      \stepcounter{@affiliationcounter}%
      \newcounter{@affil#1}%
      \setcounter{@affil#1}{\value{@affiliationcounter}}%
    \fi
  \fi%
  \ifcsname @icmlsymbol#1\endcsname
    \textsuperscript{\csname @icmlsymbol#1\endcsname\,}%
  \else
    \textsuperscript{\cfAffMark{\value{@affil#1}}\,}%
  \fi}
\renewcommand{\printAffiliationsAndNotice}[1]{%
\stepcounter{@affiliationcounter}%
{\let\thefootnote\relax\footnotetext{\hspace*{-\footnotesep}\ifdefined\isaccepted #1\fi%
\forloop{@affilnum}{1}{\value{@affilnum} < \value{@affiliationcounter}}{
\textsuperscript{\cfAffMark{\value{@affilnum}}}\ifcsname @affilname\the@affilnum\endcsname%
\csname @affilname\the@affilnum\endcsname%
\else
{\bf AUTHORERR: Missing \textbackslash{}mlsysaffiliation.}
\fi
}.
\ifdefined\mlsyscorrespondingauthor@text
Correspondence to: \mlsyscorrespondingauthor@text.
\else
{\bf AUTHORERR: Missing \textbackslash{}mlsyscorrespondingauthor.}
\fi

\ \\
\Notice@String
}}}

\makeatletter
\g@addto@macro\UrlBreaks{%
  \do\a\do\b\do\c\do\d\do\e\do\f\do\g\do\h\do\i\do\j\do\k\do\l\do\m
  \do\n\do\o\do\p\do\q\do\r\do\s\do\t\do\u\do\v\do\w\do\x\do\y\do\z
  \do\A\do\B\do\C\do\D\do\E\do\F\do\G\do\H\do\I\do\J\do\K\do\L\do\M
  \do\N\do\O\do\P\do\Q\do\R\do\S\do\T\do\U\do\V\do\W\do\X\do\Y\do\Z
  \do\0\do\1\do\2\do\3\do\4\do\5\do\6\do\7\do\8\do\9}
\makeatother

\newcommand{\Sys}{ChainFlow\xspace}
\newcommand{\TargetFamily}{Qwen3.5\xspace}          % TODO confirm family name
\newcommand{\TargetSizes}{4B, 9B and 27B}
 
\newcommand{\SpeedupEnMath}{3.23$\times$}      % English, math, 27B, 8x4 tree (b300_tables.json scaling; 1x B300, vLLM 0.31.0)
\newcommand{\SpeedupTrFC}{2.69$\times$}        % Turkish, function calling, 27B, 8x4 tree (b300_tables.json turkish)
\newcommand{\ConcurrencyLimit}{4}              % largest concurrency with tree V>1 (tab:serving, 4B, 1x B300)
 
\newcommand{\Kdraft}{8}                        % drafted positions per pass
\newcommand{\FlowSteps}{2}                     % Euler steps during training
\newcommand{\ServeFlowSteps}{1}                % Euler steps during serving
\newcommand{\CtxLen}{8}                      % context states read by the drafter
\newcommand{\Kserve}{3}                      % served chain length (tab:main chain arm)
\newcommand{\PathOrder}{8}                   % parent tokens seen by the path head
\newcommand{\MarkovRank}{256}
\newcommand{\TopM}{8192}                     % candidate ids rescored per draft
\newcommand{\cfLR}{$5\times10^{-4}$}
\newcommand{\cfEpochs}{15}
\newcommand{\cfTODO}{\textbf{XX}\xspace}        % placeholder for a missing measurement

\begin{document}

\twocolumn[{
\mlsystitle{ChainFlow: Few-Step Latent Flow Matching for Speculative Decoding}

\mlsyssetsymbol{equal}{\textit{Q}}
\mlsyssetsymbol{corr}{*}
\renewcommand{\mlsysaffilmark}[1]{\ifcase#1\or\textit{K}\or\textit{V}\else#1\fi}

\begin{mlsysauthorlist}
\mlsysauthor{Selim Aktas}{equal,ku}
\mlsysauthor{Yusuf Erdem Erdem}{equal,ytu}
\mlsysauthor{Ahmed Zeer}{corr,equal,ytu}
\mlsysauthor{Himmet Toprak Kesgin}{ytu} \\
\mlsysauthor{Muzaffer Kaan Yuce}{ytu}
\mlsysauthor{Mehmet Fatih Amasyali}{ytu}
\end{mlsysauthorlist}
\mlsysaffiliation{ku}{Department of Electronics Engineering, Koç University, Istanbul, Türkiye}
\mlsysaffiliation{ytu}{Department of Computer Engineering, Yıldız Technical University, Istanbul, Türkiye}
\mlsyscorrespondingauthor{Ahmed Zeer}{ahmed.zeer@std.yildiz.edu.tr}

\vspace{0.4em}

\begin{center}
\includegraphics[width=\textwidth]{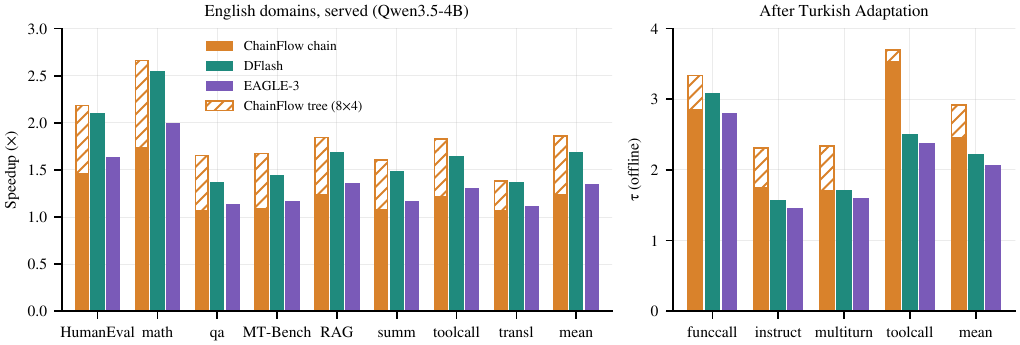}
\captionof{figure}{ChainFlow against EAGLE-3 and DFlash. Hatched: gain from 8$\times$4 tree verification. (a)~Served speedup over autoregressive decoding at batch one on English domains, ran on one NVIDIA B300 with vLLM 0.31.0 and greedy decoding. The tree beats both in every domain. (b)~Acceptance length $\tau$ (offline) after Turkish adaptation; on average even the chain beats both, showing adaptability.}
\label{fig:teaser}
\end{center}

\vskip 0.3in
\begin{abstract}
Speculative decoding has become a standard way to accelerate LLM inference: a lightweight drafter proposes tokens that the target verifies in one parallel pass. Existing drafters face trade-offs that limits their speedups: autoregressive drafters generate tokens one at a time, so drafting cost grows with draft length, while diffusion drafters generate tokens in parallel but predict each position independently, ignoring inter-token dependencies. We introduce ChainFlow, a new speculative decoding paradigm and the first flow-matching drafter conditioned on the target's hidden states. ChainFlow treats drafting as a short transport over hidden states, needing one or two flow steps instead of the many denoising steps of generation from noise. A variational autoencoder compresses these states so drafting cost barely grows with target size, lightweight feedback heads restore inter-token dependencies, and tree drafting raises acceptance. On Qwen3.5 models, at matched drafter size and training data, ChainFlow outperforms EAGLE-3 and DFlash in vLLM by 39\% and 11\%, respectively. These gains further widen as drafters are adapted to new languages sequentially. Our ablations cover drafter design and concurrent serving, and we open-source our code and drafter models.
\end{abstract}
}]

\printAffiliationsAndNotice{}

\section{Introduction}
% =====================================================================

Large language models now sit behind a broad range of deployed systems, and the cost of running them is dominated by generation. Decoding is sequential: each token requires a full pass over the model's parameters, so inference is bound by memory bandwidth rather than compute~\citep{shazeer2019mqa}. Long reasoning traces and agentic workloads sharpen the problem, because output length now sets latency.

Speculative decoding breaks part of this serialisation~\citep{leviathan2023speculative,chen2023accelerating}. A cheap drafter proposes future tokens; the target verifies them in one parallel pass and accepts the longest prefix consistent with its own distribution, leaving the output distribution unchanged. Per-token latency is then $L = (T_{\mathrm{draft}} + T_{\mathrm{verify}})/\tau$ for $\tau$ accepted tokens per cycle, and drafters sit at two poles of that trade-off. Autoregressive drafters such as EAGLE-3~\citep{li2025eagle3} condition each token on the last and reach high $\tau$, but pay $T_{\mathrm{draft}} \propto \gamma$ and must stay shallow; drafting is reported to consume 20--40\% of inference time~\citep{liu2026dart}. Parallel drafters~\citep{chen2026dflash, cheng2026dspark} emit a whole block in one pass and can afford depth, but predict every position from the verified prefix alone. Their positions are marginal, so dependence between drafted tokens has to be restored after the fact.

Autoregressive drafters buy coherence with sequential cost; parallel drafters buy speed, but at a price we question. Each round begins from an uninformative seed, mask embeddings or noise, so the drafter rebuilds the block from scratch, and coherence is patched back afterwards by $n$-gram scores~\citep{liu2026dart}, recurrent correction heads~\citep{cheng2026dspark}, or parent-conditioned trees~\citep{ringel2026ddtree}. Yet the target has already indicated where its next states lie, in its own recent hidden trajectory. A drafter seeded from that trajectory need not generate the block; it only has to \emph{transport} an estimate onto the target's future states. Flow matching fits this view, since it constrains neither the source distribution nor the number of integration steps, and short transports need few~\citep{lipman2023flowmatching,lipman2024fmguide}.

We introduce \Sys, a drafter that treats drafting as short-horizon transport on the target's hidden-state manifold. A variational autoencoder compresses the target's final hidden states to a latent space, and a conditional flow network transports an estimate of the next $\Kdraft$ latents, taken from the recent context, onto the target's future latents in one Euler step at serving time. Context states enter as cross-attention memory, so the drafter reads the target's internal representation without being embedded in it~\citep{samragh2025future}. Decoded latents are projected by the target's frozen language modeling head (lm\_head), so the drafter emits exact per-position categoricals and standard verification applies. A low-rank Markov bias and a path-conditioned residual then restore dependence on the sampled prefix without a second flow pass, and a beam search emits the drafts as a chain or a draft tree. One forward pass yields all $\Kdraft$ positions, and the flow never touches the backbone, so the drafter runs on an unmodified engine.

% src: B300 served results; see Tables tab:main and tab:scaling.
We train drafters for the \TargetSizes{} models of \TargetFamily family. At batch one, \Sys reaches \textbf{\SpeedupEnMath{}} on English mathematics and \textbf{\SpeedupTrFC{}} on Turkish function calling at 27B (English per-domain results in Table~\ref{tab:domains}). Drafters are trained first on English, then adapted on Turkish with data covering math, code, chat and open-ended sources. The English-only drafter is a net loss on Turkish; after adaptation Turkish improves sharply while English retains a speedup, so a drafter can be carried to a new language without retraining from scratch. German and French are then added in sequence, and replaying a small share of Turkish limits forgetting. Figure~\ref{fig:teaser} summarises the comparison: served speedup on English domains (a), and acceptance length after Turkish adaptation (b). A context-derived source also makes the transport short: it trains faster than noise and stays accurate with a single integration step, where noise collapses, and one step is the fastest served configuration (\S\ref{sec:design}). We also ladder concurrency: as with any speculative decoder, the gain narrows as the decode batch grows, and on one NVIDIA B300 the tree gains up to a batch of \ConcurrencyLimit{}.

Our contributions are as follows.
\begin{itemize}
\item \textbf{Faster and more adaptable.} ChainFlow is 1.39$\times$ faster than
EAGLE-3 and 1.11$\times$ faster than DFlash on English at matched drafter
size, with the highest pooled speedup at 27B and the largest adaptation gains in
Turkish, German and French.
\item \textbf{Latent transport drafter.} Flow matching over a VAE latent
of the target's hidden states, started from the extrapolated trajectory,
drafts all positions in one or two steps.
\item \textbf{Tree drafting at chain cost.} One pass yields every
position's marginals, and token feedback turns them into chains or trees
without re-running the drafter.
\end{itemize}

% =====================================================================
\section{Related Work}
% =====================================================================

\paragraph{Speculative decoding.} Blockwise parallel decoding first proposed several positions against a single verifier pass~\citep{stern2018blockwise}, and rejection-sampling verification extended this to non-greedy generation while preserving the target distribution~\citep{leviathan2023speculative,chen2023accelerating}. Tree verification packs many candidate continuations into one attention pattern, so a rejected sibling does not invalidate a whole branch~\citep{miao2024specinfer}, and later work shapes that tree from drafter confidence~\citep{li2024eagle2}. Training-free drafters built on prompt lookup or $n$-gram retrieval occupy the opposite extreme: drafting is nearly free, $\tau$ is low, and end-to-end gains stay modest~\citep{saxena2023pld,fu2024lookahead}. A separate line treats the verification budget as a scheduling problem, pruning low-confidence proposals or adapting draft length to engine load and confidence~\citep{cheng2026dspark,sadhukhan2025magicdec}. \Sys inherits these, and our experiments include a decode-batch cutoff.

\paragraph{Evolution of drafter architectures.}
Early studies use a smaller model from the target's family, whose generation latency dominates the inference time \citep{leviathan2023speculative,chen2023accelerating}. Medusa~\citep{cai2024medusa} and Hydra~\citep{ankner2024hydra} remove that second model with parameter-efficient heads on top-layer features, and the EAGLE line~\citep{li2024eagle,li2025eagle3} raises $\tau$ by predicting the next feature and reusing the target's head. While effective, these methods remain sequential in draft length. Block-parallel and diffusion drafters~\citep{an2025pard,chen2026dflash, liu2026dart} remove that dependence, permitting deeper drafters at constant latency, and DiffuSpec~\citep{li2025diffuspec} shows the same with a pre-trained diffusion LM as the drafter. Because their per-position distributions are marginal, a further family restores intra-block dependency with lightweight sequential heads~\citep{cheng2026dspark,huang2026domino} or branch-conditioned trees~\citep{hu2026jetspec,ringel2026ddtree}.

\paragraph{Flow matching for sequence generation.}Flow matching learns a velocity field transporting a source distribution to a data distribution along prescribed paths, without requiring an uninformative source~\citep{lipman2023flowmatching}. Its discrete counterpart~\citep{gat2024dfm,campbell2024discreteflows} generalises masked diffusion via wider paths and couplings, placing diffusion language models~\citep{nie2025llada} and block-diffusion drafters~\citep{arriola2025blockdiffusion} in a larger design space. Two recent results establish important properties: flow map language models~\citep{lee2026flowmap} learn the endpoint map directly, generating in one or two steps rather than integrating a long trajectory, and LangFlow~\citep{chen2026langflow} shows continuous flows over token embeddings match discrete formulations at comparable step budgets. Flow matching has also been proposed to replace or support autoregression, as a discrete or continuous flow over the full sequence~\citep{wang2025fudoki,shao2025calm}; conversely, speculation accelerates sampling from flow models~\citep{flowcast2026}. No prior work uses a flow-matching transport as an autoregressive drafter, or conditions one on the target's hidden states, which, as we show, enables efficient few-step lossless drafting.

% =====================================================================
%  ChainFlow -- Section 3 (Methodology) and Section 4 (Experimental Results)
%  SKELETON  only currently. Each paragraph slot carries keyword reminders as a
%  comment; no prose is written yet.
%
%  Convention:
%    \paragraph{...}  = an experiment / setup item, as in the plan
%    % >>              = keyword reminder for what goes in that slot
%    % CUT:            = what to drop or move to appendix if over length
% =====================================================================

% =====================================================================
\section{Methodology}
\label{sec:method}
% =====================================================================

\begin{figure*}[!t]
\centering
\includegraphics[width=\textwidth]{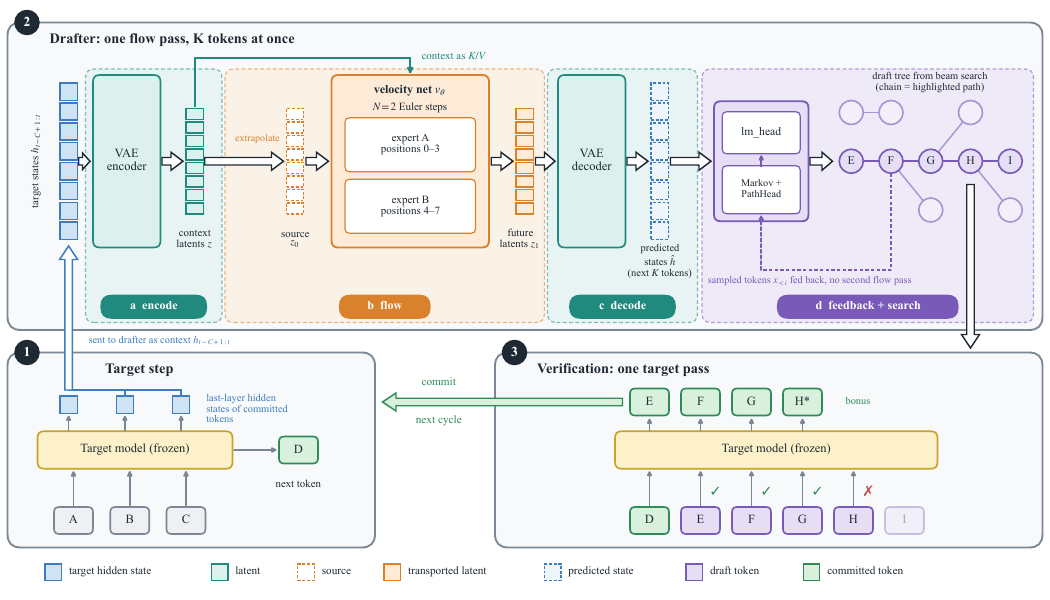}
\caption{One draft--verify cycle of \Sys on a letter example. The drafter
starts each position from the extrapolation
$z_0^{i}=z_C+i\,(z_C-z_{C-1})$ of the context latents. The illustration uses $N{=}2$ Euler steps;
served experiments use $N{=}1$. Here $E,F,G$ are accepted and $H$ is replaced by the target's own
token $H^{*}$.}
\label{fig:arch}
\end{figure*}

\Sys follows the standard draft--verify cycle: the drafter proposes a block
of continuations, the target scores all of them in one forward pass, and the
longest accepted prefix is committed together with the target's own bonus
token. Throughout, $C$ denotes the number of context states the drafter
reads, $K$ the number of positions it drafts and $N$ the number of
integration steps; we use $C=\CtxLen$ and $K=\Kdraft$, train with
$N=\FlowSteps$, and serve with $N=\ServeFlowSteps$. The two differ by design: we train with $N=2$ and ablate how many steps the trained transport needs at inference, finding that a single step suffices and is fastest when served (\S\ref{sec:design}).
Two problems shape the design. A block-parallel drafter produces all
positions at once, so each position is a marginal over its predecessors and
acceptance decays along the block; \S\ref{sec:transport} addresses this with
latent transport and post-hoc token feedback. And a longer block amortises
verification only if the extra candidates are worth proposing and verifying;
\S\ref{sec:candidates} covers candidate search and verification.
Figure~\ref{fig:arch} walks through one cycle.

% Figure source: figs/make_arch.py (matplotlib, vector PDF).

% ---------------------------------------------------------------------
\subsection{Latent Transport with Token Feedback}
\label{sec:transport}
% ---------------------------------------------------------------------

\Sys pays a fixed number of network evaluations for the whole block, but a
block-parallel prediction is a marginal: position $i$ cannot condition on
the token sampled at $i-1$, so acceptance decays along the block whenever
the context admits several continuations~\citep{cheng2026dspark}. We attack
both ends of this. The marginals themselves improve because the flow
transports from an informative source rather than noise; the conditionals
are restored after the flow by corrections that cost almost nothing.

\paragraph{Hidden-state compression.}
Following latent diffusion~\citep{rombach2022latentdiffusion} and latent
flow matching~\citep{dao2023latentflow,esser2024rectifiedflow}, we use a VAE
to separate representation compression from future-state modeling. Its
bottleneck reduces the dimensionality of the flow's prediction problem,
while the reconstruction objective encourages retention of information
needed to recover the target's hidden states.
A transformer VAE maps sequences of target hidden states to a
lower-dimensional latent, $z=\mathrm{Enc}_\phi(h)$ and
$\hat h=\mathrm{Dec}_\psi(z)$. Encoder and decoder are two-layer
bidirectional transformers over windows of up to 16 states, so each latent
also reflects its neighbours; dimensions are in Table~\ref{tab:models}.
Because the flow's cost scales with the latent width $Z$ rather than $D$,
one network design serves every target size. The VAE is trained on cached
hidden states with a small KL weight ($\beta=3\times10^{-5}$), which
discourages arbitrarily scaled latent representations while allowing
faithful reconstruction~\citep{rombach2022latentdiffusion}. It is then
fine-tuned jointly with the drafter under a reconstruction term
(\S\ref{sec:training}); the drafter uses the posterior mean, so encoding is
deterministic.

\paragraph{Flow matching with trajectory extrapolation.}
Let $c=(z_{t-C+1},\dots,z_t)$ be encoded context and
$z_1\in\mathbb{R}^{K\times Z}$ encoded future states. We use the
rectified-flow path $z_s=(1-s)\,z_0+s\,z_1$ for
$s\in[0,1]$~\citep{lipman2023flowmatching}, training a velocity network
$v_\theta$ to regress its constant velocity:
\begin{equation}
\mathcal{L}_{\mathrm{flow}}
 = \mathbb{E}_{s}\,\big\| v_\theta(z_s, s, c) - (z_1 - z_0) \big\|^2 .
\label{eq:flow}
\end{equation}
Flow matching places no constraint on the source $z_0$, and we exploit this.
Instead of Gaussian noise, draft position $i\in\{1,\dots,K\}$ starts from a
linear extrapolation of the last two context latents,
\begin{equation}
z_0^{(i)} = z_t + i\,(z_t - z_{t-1}),
\label{eq:source}
\end{equation}
which we call \emph{trajectory extrapolation}. Future
states lie close to the recent trajectory, so the transport is short and
nearly straight, and one or two Euler steps
$z\leftarrow z+\tfrac{1}{N}\,v_\theta(z,s,c)$ suffice. Repeating the last
latent and pure noise are the alternatives compared in \S\ref{sec:design}.

\paragraph{Velocity network.}
$v_\theta$ is a pair of \emph{chunk experts}: two separately parameterised
networks, each responsible for one chunk of four consecutive draft positions
(there is no routing; every expert always runs). Each expert is a stack of 8
pre-norm blocks with 8 heads and learned position and time embeddings. A block
applies causal self-attention over the draft positions, cross-attention to the
context latents $c$ as keys and values, and a feed-forward layer of $6\times$
width. The context enters only as keys and values, so the drafter reads the
target's representation without being embedded in it~\citep{samragh2025future}.
The two experts run concurrently, and a draft costs $N$ network evaluations
regardless of $K$. We ablate the number of experts in \S\ref{sec:design}.

\paragraph{Post-hoc token feedback.}
After the last Euler step the decoder returns one hidden state per drafted
position, $\hat h_{1:K}=\mathrm{Dec}_\psi(z)$, and a single matrix product
with the frozen \texttt{lm\_head} $W$ turns all $K$ of them into logits.
Two lightweight corrections restore dependence on the chosen prefix after
the flow:
% Stacked hboxes instead of split/aligned: the review-mode line numbering
% swallowed the \\ breaks, leaving one overlong line.
\begin{equation}
\vcenter{%
\hbox{$\displaystyle p_\theta(x_i \mid x_{<i}) = \mathrm{softmax}\Big(W\big[\hat h_i + r_\omega(x_{i-p:i-1})\big]$}%
\hbox{$\displaystyle \qquad\qquad\qquad {}+ U_2\,U_1[x_{i-1}]\Big).$}}
\label{eq:feedback}
\end{equation}
The path head $r_\omega$ sums per-offset projections of the frozen token
embeddings of the last $p=\PathOrder$ parent tokens and passes the sum
through a small MLP with a zero-initialised output layer, so training starts
from the marginal flow; it corrects the hidden state before the
\texttt{lm\_head}. The second term is a Markov head, as
in~\citet{cheng2026dspark}: a rank-\MarkovRank{} factorised first-order
transition bias, $U_2\,U_1[x_{i-1}]$, that conditions each position on the
immediately preceding sampled token. Neither term re-runs the flow or the
VAE; only these corrections are evaluated per position. Training uses the
target's tokens as parents; inference uses the tokens chosen along each
beam.

% ---------------------------------------------------------------------
\subsection{Candidate Search and Verification}
\label{sec:candidates}
% ---------------------------------------------------------------------

A longer draft block amortises the fixed verification cost over more
accepted tokens, but only if the extra candidates are worth proposing and
verifying. This subsection decides what to propose, how to search it, and how
it is verified.

\paragraph{Two-pass candidate head.}
Scoring every candidate against the full \texttt{lm\_head} would dominate
draft time, so the head is restricted to a shortlist of the 62{,}642 tokens
that occur in the training corpus. One matrix product of $\hat h_{1:K}$
against the shortlist selects a shared set of $M=\TopM$ candidate ids, and
Eq.~\ref{eq:feedback} is evaluated only over those rows. This affects only
what the drafter can propose: a missing token lowers acceptance but cannot
change the target's output.

\begin{figure}[h]
  \centering
  \includegraphics[width=0.8\columnwidth]{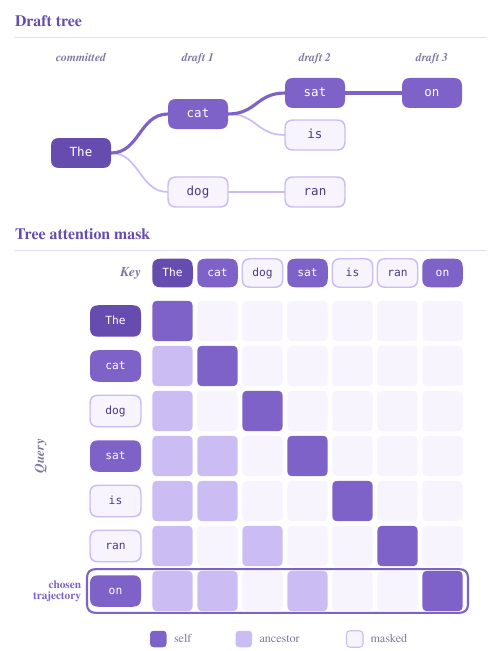}
  \caption{Tree drafting and attention mask (illustrative). Each draft token attends only to itself and its ancestors, so all branches are verified in one target pass.}
  \label{fig:tree-attn}
\end{figure}

\paragraph{Chain and tree emission.}
A beam search over Eq.~\ref{eq:feedback} emits the candidates as a chain or a tree (Figure~\ref{fig:tree-attn}). A chain keeps only the best path, of length \Kserve{}. A tree of shape $W_b\times D$ is built level by level: at each depth, the nodes of the previous level are expanded with their top candidate tokens, all resulting children are ranked by cumulative log-probability, and the best $W_b$ are kept, each with a pointer to its parent. Our headline $8\times4$ tree ($W_b{=}8$, $D{=}4$) therefore has 32 draft nodes, and a rejected sibling does not invalidate the round. Tree shape trades verification cost against acceptance, so we vary $W_b$ and $D$ in an ablation (\S\ref{sec:design}). The target verifies a tree in one pass by flattening its nodes into a sequence under an ancestor mask: each node attends only to the committed prefix and its ancestors, so every root-to-node path is scored as a separate chain. Chains are verified by standard rejection; tree verification needs forked recurrent state for \TargetFamily's linear-attention layers and supports greedy decoding only. In both modes a token is emitted only if the target's own logits select it, up to fp16 ties.

% ---------------------------------------------------------------------
\subsection{Training}
\label{sec:training}
% ---------------------------------------------------------------------

\paragraph{Data.}
The target generates a greedy response for every training
prompt~\citep{li2025eagle3,chen2026dflash}, and one forward pass caches its
final hidden states. Each example is a window anchored at a response
position $t$: the states $h_{t-C+1:t}$ form the context, and the next $K$
hidden states and tokens $x^\star_{1:K}$ are the targets.

\paragraph{Objective.}
The flow, path head, Markov head and VAE are trained end-to-end with
\begin{equation}
\vcenter{%
\hbox{$\displaystyle \mathcal{L} = 0.2\,\mathcal{L}_{\mathrm{flow}}
 + 0.25\,\mathcal{L}_{\mathrm{hid}} + 0.3\,\mathcal{L}_{\mathrm{CE}}$}%
\hbox{$\displaystyle \quad {}+ 0.3\,\mathcal{L}_{\mathrm{cov}}
 + 0.15\,\mathcal{L}_{\mathrm{acc}} + 0.5\,\mathcal{L}_{\mathrm{rec}}.$}}
\label{eq:loss}
\end{equation}
$\mathcal{L}_{\mathrm{hid}}$ is the relative squared error against the
target's hidden states, with and without the path correction, evaluated on
predictions integrated through the full $N$-step trajectory so it sees the
trajectory used at inference. $\mathcal{L}_{\mathrm{CE}}$ is the
cross-entropy of Eq.~\ref{eq:feedback} plus a $0.2\times$ term on the
marginal logits, which keeps the uncorrected flow decodable.
$\mathcal{L}_{\mathrm{cov}}$ is a hinge that pushes the true token into the
top $b=8$ logits, preparing the drafter for tree expansion.
$\mathcal{L}_{\mathrm{rec}}$ is the VAE reconstruction error. Finally,
\begin{equation}
\mathcal{L}_{\mathrm{acc}} = -\sum_i \gamma^{i}
 \prod_{j\le i} p_\theta(x^\star_j \mid x^\star_{<j})
\label{eq:acc}
\end{equation}
with $\gamma=0.8$ is a differentiable surrogate for the expected accepted
length.

\paragraph{Optimization.}
All modules train with AdamW at a constant \cfLR{} for \cfEpochs{} epochs,
with batches of about 6K windows and the VAE at $1/20$ of the drafter's
learning rate. Adaptation to a new language warm-starts from the English
drafter and continues with the same objective at $1$--$1.5\times10^{-4}$
for 2--6 epochs. Languages are added in sequence (Turkish, then German, then
French), optionally mixing in a share of earlier-language rows as replay
(\S\ref{sec:multilingual}).

\section{Experiments}
\label{sec:experiments}
% =====================================================================
% Order: setup -> main results -> multilingual -> drafter design ->
% draft cost and serving. "\cfTODO" = missing measurement; every number
% carries a "% src:" comment. Text states trends; numbers live in tables.
% =====================================================================

\newcommand{\cfParamsS}{529M}               % src: report p.9 (529,420,800)
\newcommand{\cfParamsM}{1.16B}              % src: report p.14 (1,156,534,272)
\newcommand{\cfParamsL}{1.64B}              % src: report p.14 (1,642,131,456)
\newcommand{\DevTarget}{\cfTODO}            % TODO target of the development sweeps (0.8B?)
\newcommand{\DevK}{\cfTODO}                 % TODO drafted positions in the development sweeps
\newcommand{\DevData}{\cfTODO}              % TODO data of the development sweeps (GSM8K?)

% ---------------------------------------------------------------------
\subsection{Experimental Setup}
\label{sec:expsetup}
% ---------------------------------------------------------------------

\paragraph{Target and draft models.}
We train one drafter for each \TargetFamily{}~\citep{qwen35} target
(Table~\ref{tab:models}). All drafters read $C=\CtxLen$ context states,
draft $K=\Kdraft$ positions, are trained with $N=\FlowSteps$ Euler steps, and use 8 blocks
with an FFN multiplier of 6; only the latent width changes with the target.
Drafts are served as chains of \Kserve{} tokens or as $8\times4$ trees,
with a single Euler step (\S\ref{sec:design}).
% src: tab:main arm.json (KEEP/DEPTH, STEPS=1).

\begin{table*}[!t]
\centering\small
\setlength{\tabcolsep}{2.5pt}
\caption{English results at \TargetFamily-4B, batch one: acceptance length $\tau$ and speedup $V$ over autoregressive decoding by domain and pooled mean. C denotes chain drafting ($K{=}\Kserve$); T1 and T2 denote $8\times4$ and $5\times7$ trees (nodes per depth $\times$ depth), respectively, with T1 as our headline configuration. Each drafter has $\sim$0.52B parameters. EAGLE-3 and DFlash draft $K{=}4$ tokens per round. Evaluated on NVIDIA B300 with vLLM~0.31.0 and greedy decoding.}
\label{tab:main}
\vskip 0.1in
\begin{tabular}{l *{8}{cc} | cc}
\toprule
Method & \multicolumn{2}{c}{Code} & \multicolumn{2}{c}{Math} & \multicolumn{2}{c}{QA} & \multicolumn{2}{c}{MT} & \multicolumn{2}{c}{RAG} & \multicolumn{2}{c}{Summ.} & \multicolumn{2}{c}{Tool} & \multicolumn{2}{c}{Transl.} & \multicolumn{2}{c}{Mean} \\
\cmidrule(lr){2-3} \cmidrule(lr){4-5} \cmidrule(lr){6-7} \cmidrule(lr){8-9} \cmidrule(lr){10-11} \cmidrule(lr){12-13} \cmidrule(lr){14-15} \cmidrule(lr){16-17} \cmidrule(lr){18-19}
& $\tau$ & $V$ & $\tau$ & $V$ & $\tau$ & $V$ & $\tau$ & $V$ & $\tau$ & $V$ & $\tau$ & $V$ & $\tau$ & $V$ & $\tau$ & $V$ & $\tau$ & $V$ \\
\midrule
EAGLE-3 & 3.09 & 1.64$\times$ & 3.80 & 2.00$\times$ & 2.09 & 1.14$\times$ & 2.12 & 1.17$\times$ & 2.48 & 1.36$\times$ & 2.10 & 1.17$\times$ & 2.41 & 1.31$\times$ & 2.30 & 1.11$\times$ & 2.49 & 1.35$\times$ \\
DFlash  & 3.02 & 2.11$\times$ & 3.68 & 2.55$\times$ & 1.89 & 1.38$\times$ & 1.98 & 1.44$\times$ & 2.34 & 1.69$\times$ & 2.03 & 1.49$\times$ & 2.29 & 1.64$\times$ & 2.24 & 1.38$\times$ & 2.35 & 1.69$\times$ \\
\midrule
\Sys{} C  & 2.43 & 1.45$\times$ & 2.91 & 1.73$\times$ & 1.72 & 1.06$\times$ & 1.76 & 1.08$\times$ & 1.99 & 1.23$\times$ & 1.72 & 1.07$\times$ & 1.98 & 1.21$\times$ & 1.85 & 1.06$\times$ & 2.01 & 1.23$\times$ \\
\Sys{} T1 & 3.31 & \textbf{2.19$\times$} & 4.07 & 2.67$\times$ & \textbf{2.43} & \textbf{1.66$\times$} & \textbf{2.44} & \textbf{1.68$\times$} & \textbf{2.73} & \textbf{1.85$\times$} & \textbf{2.32} & \textbf{1.61$\times$} & \textbf{2.74} & \textbf{1.84$\times$} & \textbf{2.58} & \textbf{1.39$\times$} & \textbf{2.78} & \textbf{1.87$\times$} \\
\Sys{} T2 & \textbf{3.41} & 2.14$\times$ & \textbf{4.51} & \textbf{2.78$\times$} & 2.31 & 1.51$\times$ & 2.33 & 1.53$\times$ & 2.65 & 1.72$\times$ & 2.23 & 1.48$\times$ & 2.64 & 1.69$\times$ & 2.46 & 1.32$\times$ & 2.73 & 1.75$\times$ \\
\bottomrule
\end{tabular}
\end{table*}

\begin{table}[!htbp]
\centering\small
\setlength{\tabcolsep}{5pt}
\caption{Drafters. Parameter counts exclude the target's shared, frozen
embedding and \texttt{lm\_head}.}
\label{tab:models}
\vskip 0.1in
\begin{tabular}{lccc}
\toprule
Target & Hidden $D$ & Latent $Z$ & Drafter params \\
\midrule
\TargetFamily-4B  & 2560 & 640  & \cfParamsS \\
\TargetFamily-9B  & 4096 & 1024 & \cfParamsM \\
\TargetFamily-27B & 5120 & 1024 & \cfParamsL \\
\bottomrule
\end{tabular}

\end{table}

\paragraph{Training data.}
Drafters are first trained on 45.9K English prompts covering mathematics and
science, code, dialogue, instruction following, summarisation, translation
and creative writing. They are then adapted in sequence, without English
replay, to Turkish, German and French on domains covering instruction following, function and tool calling,
multi-turn dialogue and retrieval-augmented generation. Training responses are the
target's own greedy continuations. Appendix~\ref{sec:appendix-data} lists
every source.

\paragraph{Benchmarks.}
English uses the eight RedHatAI speculator benchmark domains: code (HumanEval~\citep{chen2021humaneval}), mathematical reasoning, question answering, MT-Bench, retrieval-augmented generation, summarisation, tool calling and translation. Other languages use held-out prompts of their training sources. Generations are limited to 256 tokens, and tokens are pooled across domains.

\paragraph{Protocol and metrics.}
All methods run on one NVIDIA B300 GPU with vLLM 0.31.0~\citep{kwon2023vllm}, greedy decoding and batch size one unless stated, with asynchronous scheduling for every arm including the baseline. Requests stop at their natural end of sequence or at 256 new tokens, and each served number pools two or three server launches. The \Sys{} drafter is implemented as a custom proposer in vLLM.

We report \emph{speedup} in output tokens per second over autoregressive decoding and
\emph{acceptance length} $\tau$, the tokens produced per round including
the bonus token. Cross-method multilingual and per-position comparisons use
$\tau$ measured offline on cached target states, with each drafter given
the information the engine gives it when serving.
% src: BENCHMARKING.md sec.3; report p.2-3 (plugin mode, 600 windows/domain).
% src: vLLM build read from each server.log by vllm/paper_fill.py (0.31.0 for every arm); tab:main 3 launches per arm, everything else 2. Runner control: b1 AR on V1 = f_base_v1 (0/200 texts differ from V2).

\paragraph{Comparisons.}
We compare against EAGLE-3~\citep{li2025eagle3} and
DFlash~\citep{chen2026dflash}, retrained on the same data with drafters
matched in size to ours at 4B and served on the same engine, drafting
$K{=}4$ tokens per round, the depth of our headline tree.
We additionally evaluate scaling across the 4B, 9B, and 27B targets
(Table~\ref{tab:scaling}) and examine drafter design choices through
development sweeps (\S\ref{sec:design}). These cover token feedback, expert
partitioning, source initialization, integration steps, network capacity,
and draft tree construction.

% ---------------------------------------------------------------------
\subsection{Main Results}
\label{sec:main}
% ---------------------------------------------------------------------

\paragraph{Comparisons.}
Pooled over the eight domains, the \Sys{} $8\times4$ tree is 1.11$\times$ faster than DFlash and 1.39$\times$ faster than EAGLE-3 (Table~\ref{tab:main}), and it leads DFlash in every domain, by the least on translation (1.39$\times$ against 1.38$\times$). The chain, which needs no engine change, reaches 1.23$\times$ and trails both baselines; tree verification lifts $\tau$ from 2.01 to 2.78.
Gains follow how predictable a domain is: code and mathematics, whose
outputs are structured and partly copied from the prompt, gain most, and
translation, whose output language differs from the context, least.

\begin{table}[!htbp]
\centering\small
\setlength{\tabcolsep}{3.8pt}
\caption{Scaling with target size. Flow and AR latencies
are per draft and decoding step (ms), respectively.
$\tau$ is measured on a fixed $8\times4$ tree;
chain uses $K=\Kserve$.}
\label{tab:scaling}
\vskip 0.1in
\begin{tabular}{lccccc}
\toprule
Target & Flow ms & AR step ms & $\tau$ & Chain & Tree \\
\midrule
4B  & 0.30 & 3.1 & 2.78 & 1.23$\times$ & 1.87$\times$ \\
9B  & 0.38 & 4.2 & 2.66 & 1.30$\times$ & 1.78$\times$ \\
27B & 0.38 & 11.2 & 2.73 & 1.61$\times$ & 2.12$\times$ \\
\bottomrule
\end{tabular}

% src: flow ms results/flow_ms.json (fused kernel, B=1, 1 step); AR step = 1000 / AR tok/s (327.1 / 236.9 / 89.5); results/b300_tables.json scaling.
% Launches base/chain/tree: 4B 3/3/3, 9B 2/2/2, 27B 2/2/2. Chain tau 4B/9B/27B = 2.01/2.01/2.07.

\end{table}

\paragraph{Scaling with target size.}
Tree acceptance stays between 2.66 and 2.78 from 4B to 27B, while flow
integration takes 0.30--0.38\,ms and the autoregressive step grows from
3.1 to 11.2\,ms. Chain speedup rises from 1.23$\times$ to 1.30$\times$
and 1.61$\times$; tree speedup is 1.87$\times$, 1.78$\times$ and
2.12$\times$, peaking at 27B
(Table~\ref{tab:scaling}). Tree drafting is
profitable on translation at every size, with speedups of 1.39$\times$,
1.42$\times$ and 1.74$\times$.

\begin{figure*}[!t]
\centering
\includegraphics[width=0.9\textwidth]{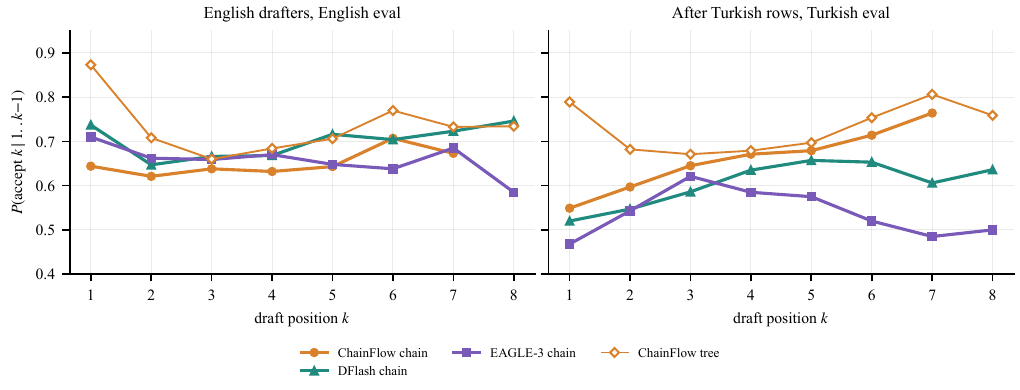}
\caption{Conditional acceptance at each draft position, given that all
earlier positions were accepted (offline, 4B). Left: English drafters on
English. Right: after Turkish adaptation, on Turkish. Chain curves end at
position 7; the tree is $8{\times}4$.}
\label{fig:position}
\end{figure*}

Table~\ref{tab:domains} breaks the tree speedup down by domain. The ordering of domains is the same at every size, with math and code highest and translation lowest. Speedup in 27B target leads in every domain, peaking at 3.23$\times$ on math.

\begin{table}[!htbp]
\centering\small
\setlength{\tabcolsep}{2.5pt}
\caption{Per-domain tree speedup ($8\times4$) by target size in English.}
\label{tab:domains}
\vskip 0.1in
\begin{tabular}{lcccccccc}
\toprule
Target & Code & Math & QA & MT & RAG & Summ. & Tool & Transl. \\
\midrule
4B  & 2.19 & 2.67 & 1.66 & 1.68 & 1.85 & 1.61 & 1.84 & 1.39 \\
9B  & 2.13 & 2.67 & 1.52 & 1.54 & 1.73 & 1.52 & 1.83 & 1.42 \\
27B & \textbf{2.51} & \textbf{3.23} & \textbf{1.82} & \textbf{1.87} & \textbf{2.09} & \textbf{1.79} & \textbf{2.09} & \textbf{1.74} \\
\bottomrule
\end{tabular}
\end{table}

\paragraph{Acceptance along the block.}
Figure~\ref{fig:position} shows the probability of accepting each draft
position given all earlier ones were accepted. Parallel drafters do not decay
along the block: acceptance is flat or rising for \Sys and DFlash, whereas
sequential EAGLE-3 declines, markedly so on Turkish. On English the chain's
curve is flat but slightly below the baselines, so its lower $\tau$ reflects a
uniformly lower per-position level, not a deeper failure. The tree keeps \Sys
at or above both baselines at nearly every position, most clearly at the first.
After Turkish adaptation both \Sys variants lead everywhere, and the chain's
acceptance rises along the block, consistent with later positions conditioning
on a longer accepted prefix.

% ---------------------------------------------------------------------
\subsection{Multilingual Adaptation}
\label{sec:multilingual}
% ---------------------------------------------------------------------

\begin{figure*}[t]
    \centering
    \includegraphics[width=\textwidth]{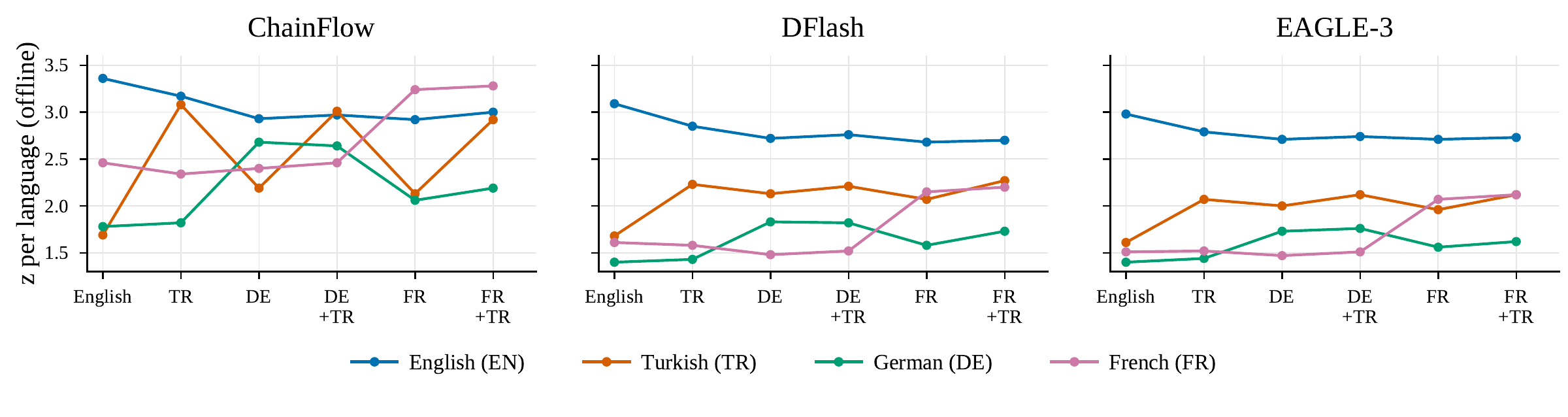}
    \caption{Effect of language adaptation on offline decoding speedup
    ($z$) for ChainFlow (tree mode), DFlash, and EAGLE-3.
    Adapting to new languages improves their decoding speedups,
    with ChainFlow achieving the largest gains, while English
    performance declines slightly across all three methods.}
    \label{fig:language_comparison}
\end{figure*}

\paragraph{Adapting to new languages.}
\Sys gains most in every new language under sequential adaptation (Figure~\ref{fig:language_comparison}),
while English performance drifts down similarly for all three methods,
so this drop does not distinguish them. Served without adaptation, an
English drafter is a net loss on Turkish at 4B and 9B; adaptation yields
tree speedups of 1.81$\times$, 1.87$\times$ and 2.25$\times$ at 4B, 9B and 27B,
respectively.

\paragraph{Retaining earlier languages.}
This plasticity has a cost. Without replay, adding German erases most of the
Turkish gain, a loss five to nine times larger than that of the baselines.
Mixing a fifth of Turkish rows into each later training stage removes nearly all
of this forgetting while leaving the new-language gain intact, so \Sys can
adapt to a new language without giving up the earlier ones.

\paragraph{Draft vocabulary.}
Part of the lead comes from the candidate head. Our per-language shortlist covers nearly all Turkish evaluation tokens, whereas the baselines' frozen English vocabulary covers only about three in five. On French, where this gap is smallest, our gain is still the largest.

\subsection{Drafter Design}
\label{sec:design}

ChainFlow combines many methods, and each component affects the end-to-end speedup differently. To understand these effects, we isolate four design decisions: token feedback, the number of experts, the source distribution, and expert width. We measure the effect of each change separately in each setting, and then combine the best-performing choices into the final drafter. All experiments use the Qwen3.5-0.8B target model, $K=4$ draft positions, and the GSM8K dataset.

\begin{table}[!htbp]
\centering\small
\setlength{\tabcolsep}{4pt}
\caption{Token feedback. acc@$i$: acceptance at draft
position $i$; speedup measured live.}
\label{tab:feedback}
\vskip 0.1in
\begin{tabular}{lcccc}
\toprule
Drafter & acc@1 & acc@4 & $\tau$ & Speedup \\
\midrule
Flow, context only          & 0.96 & 0.44 & 2.68 & 0.96$\times$ \\
\quad + deep-position loss   & 0.96 & 0.44 & 2.89 & 0.99$\times$ \\
\quad + Markov head          & 0.95 & 0.46 & 2.93 & 1.01$\times$ \\
Autoregressive & 0.99 & 0.57 & 3.21 & 1.00$\times$ \\
\bottomrule
\end{tabular}

% src: dump image 10.27.13. TODO: full-ChainFlow row in the same setting.

\end{table}

\begin{table}[!htbp]
\centering\small
\setlength{\tabcolsep}{4.5pt}
\caption{Splitting the draft block across experts.
Plateau: validation expected acceptance relative to $2\times4$.
Values are approximate.}
\label{tab:experts}
\vskip 0.1in
\begin{tabular}{lcc}
\toprule
Experts $\times$ positions & Plateau & Epochs to 90\% \\
\midrule
$2\times4$ (\Sys) & 1.00 & $\approx$85 \\
$4\times2$        & $\approx$0.99 & $\approx$80 \\
\bottomrule
\end{tabular}

% src: ablation_notes_and_figs/flow_experts.png, DIGITIZED.
% Plateaus renormalized from 0.87 and 0.86 to the two-expert result:
% 0.87/0.87 = 1.00; 0.86/0.87 = 0.9885 (approximately 0.99).

\end{table}

\begin{figure}[htbp]
    \centering
    \includegraphics[width=0.85\columnwidth]{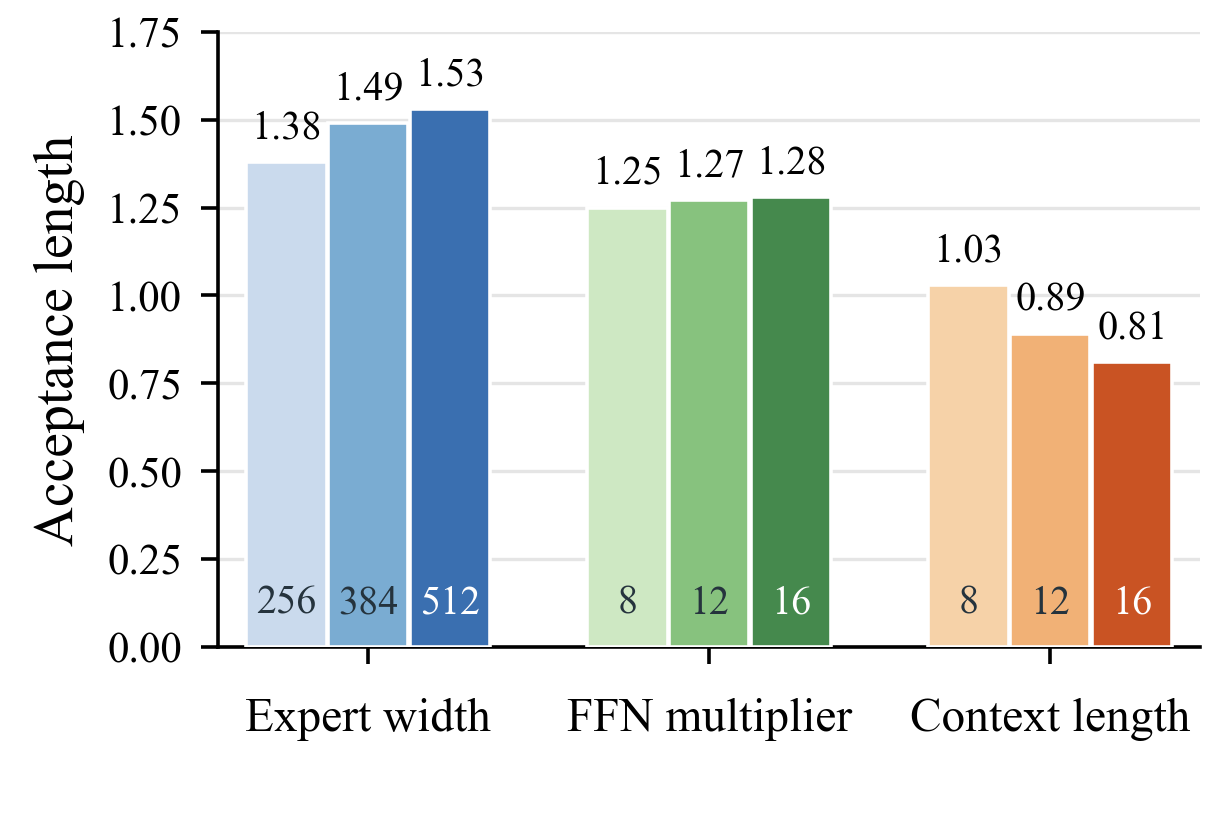}
    \caption{Capacity and context sweeps for flow with MLP experts.}
    \label{fig:capacity}
\end{figure}

\paragraph{Token feedback.}
Drafters that predict every position from the context alone accept the first
position almost always, later ones rarely, and run slower than
autoregressive decoding (Table~\ref{tab:feedback}). Loss reweighting, more
experts, other sources and more capacity all left this ceiling in place
(Tables~\ref{tab:experts} and~\ref{tab:init},
Figure~\ref{fig:capacity}). Feeding back the chosen
token lifts it: the Markov head closes part of the gap to an autoregressive
drafter within one flow pass, and the path head extends the feedback to the
full prefix and to the hidden state.

\paragraph{Experts.} Increasing the number of experts from two to four does not raise validation acceptance beyond the plateau already reached with two (Table~\ref{tab:experts}). Two experts are therefore sufficient to cover our eight-position draft, and \Sys uses two experts, each handling four positions, which avoids the cost of additional experts.

\begin{table}[!htbp]
\centering\small
\setlength{\tabcolsep}{4pt}
\caption{Source distribution. Left: training epochs to 90\% of the
validation plateau. Right: inference-time Euler steps
for 4B drafters trained at $N=2$ (offline, English chain).}
\label{tab:init}
\vskip 0.1in
\begin{tabular}{lc|ccc}
\toprule
 & Dev.\ sweep & \multicolumn{3}{c}{4B, $\tau$} \\
Source $z_0$ & epochs to 90\% & $N{=}1$ & $N{=}2$ & $N{=}4$ \\
\midrule
Noise          & $\approx$57 & 1.97 & 2.55 & 2.47 \\
Repeat last    & $\approx$38 & 2.58 & 2.66 & 2.42 \\
Extrapolation  & $\approx$20 & 2.49 & 2.53 & 2.50 \\
\bottomrule
\end{tabular}

% src: left DIGITIZED from flow_init_settings.png; right report p.4-5.

\end{table}

\begin{figure}[htbp]
\centering
\includegraphics[width=0.9\columnwidth]{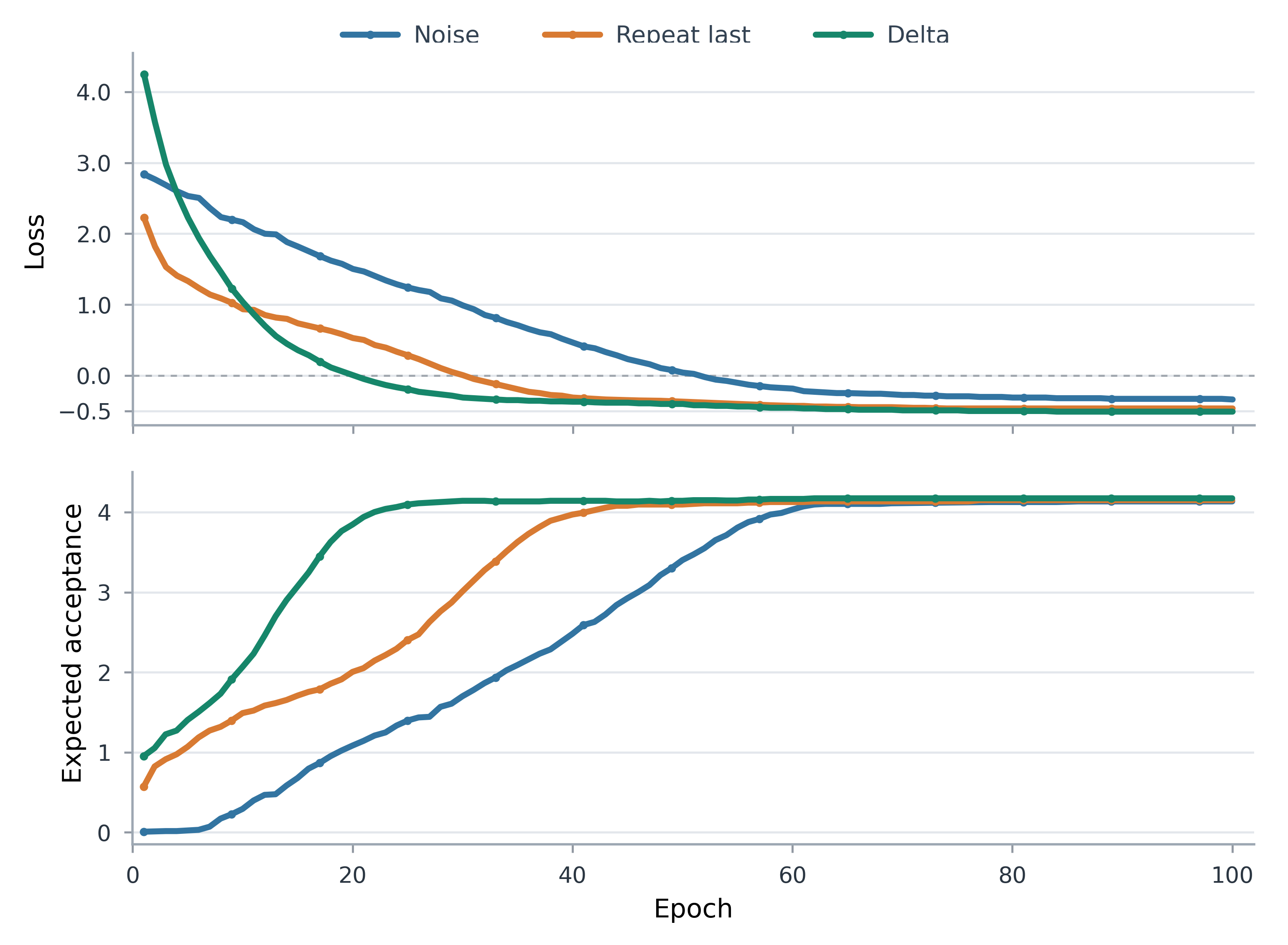}
\caption{Training loss (top) and validation expected acceptance (bottom)
across epochs for three initializations: noise, last-latent repetition,
and trajectory extrapolation (Delta).}
% src: DIGITIZED (figs/make_design_figs.py).
\label{fig:init}
\end{figure}

\paragraph{Source distribution and integration steps.}
An informative source shortens the transport. All three sources reach the
same plateau, but context-derived ones get there much faster
(Figure~\ref{fig:init}), and at inference they survive a single Euler step,
where noise collapses (Table~\ref{tab:init}). Extrapolation and repeating
the last latent do not separate at this scale.
Served, a second Euler step raises $\tau$ from 2.78 to 2.83 but reduces
throughput by 5\% (611.7 to 580.1 tokens/s), so the served configuration
uses one step.
% src: b300_tables.json ablations.steps2 (4B, 8x4 tree, batch one).

\paragraph{Capacity.}
We vary expert width, the FFN multiplier and context length in separate
development sweeps with MLP experts (Figure~\ref{fig:capacity}). Increasing
width improves acceptance, while increasing the FFN multiplier offers
little gain. Longer contexts reduce acceptance in these sweeps, motivating
our choice of $C=\CtxLen$ context states. No capacity setting lifted the
ceiling that token feedback removes.

\paragraph{Draft tree construction.}
A tree spends target compute on extra candidates, so its shape sets how acceptance trades against cost. Unlike the sweeps above, this ablation is served end to end on the 4B target at batch one: we vary the nodes kept per depth $W_b$ and the depth $D$ while holding the drafter fixed (Table~\ref{tab:tree}). Acceptance grows with both width and depth, but every extra node must be searched and verified, so throughput falls. Beyond $8\times4$, the gain in $\tau$ is smaller than the added cost, and $8\times4$ is the fastest tree. Narrow, deep trees fare worst: $5\times7$ has both the lowest $\tau$ and the lowest throughput, so depth does not pay without width. A compact tree therefore beats larger ones, and we use $8\times4$ throughout.

\begin{table}[!htbp]
\centering\small
\setlength{\tabcolsep}{4pt}
\caption{Tree shape ($W_b\times D$: nodes kept per depth $\times$ depth) at \TargetFamily-4B, batch one, eight English domains. Positions: tokens verified per round, including the committed token. Speedup is relative to plain decoding (327 tokens/s). The headline $8\times4$ tree is best.}
\label{tab:tree}
\vskip 0.1in
\begin{tabular}{lcccc}
\toprule
Tree Shape & Positions & Tokens/s & $\tau$ & Speedup \\
\midrule
$8\times4$ & 33 & \textbf{610.9} & 2.82 & \textbf{1.87$\times$} \\
$8\times5$ & 41 & 606.2 & 2.84 & 1.85$\times$ \\
$8\times6$ & 49 & 596.4 & 2.87 & 1.83$\times$ \\
$12\times4$ & 49 & 593.6 & 2.88 & 1.82$\times$ \\
$6\times5$ & 31 & 593.4 & 2.74 & 1.82$\times$ \\
$16\times4$ & 65 & 589.0 & \textbf{2.96} & 1.80$\times$ \\
$5\times7$ & 36 & 572.8 & 2.73 & 1.75$\times$ \\
\bottomrule
\end{tabular}
% src: results/tree_shape_sweep.csv (K = positions; first launch), 4B rows; plain = 327 tok/s.
\end{table}

% ---------------------------------------------------------------------
\subsection{Serving and Vocabulary Shortlisting}
\label{sec:serving}
% ---------------------------------------------------------------------

\paragraph{Serving under concurrency.}
The gain narrows as concurrency grows (Table~\ref{tab:serving}): every request sends
the tree's 33 positions through a target that is no longer memory-bound, while
acceptance stays between 2.77 and 2.85. The tree gains up to \ConcurrencyLimit{} concurrent requests, so
speculation with the tree pays off only up to that decode batch, and beyond it
plain autoregressive decoding is faster. EAGLE-3 and DFlash, which verify only $K{+}1$
positions per request, keep a gain at every level.

\begin{table}[!htbp]
\centering\small
\setlength{\tabcolsep}{2.5pt}
\caption{Speedup under concurrency at \TargetFamily-4B: output tokens per second relative to autoregressive decoding at the same number of concurrent requests.}
\label{tab:serving}
\vskip 0.1in
\begin{tabular}{lcccccc}
\toprule
 & \multicolumn{6}{c}{Concurrent requests} \\
\cmidrule(lr){2-7}
Method & 1 & 4 & 8 & 16 & 64 & 128 \\
\midrule
EAGLE-3 & 1.33 & 1.26 & 1.23 & 1.20 & 1.20 & 1.06 \\
DFlash & 1.66 & 1.48 & 1.39 & 1.33 & 1.32 & 1.14 \\
\midrule
\Sys{} & 1.85 & 1.12 & 0.94 & 0.81 & 0.63 & 0.45 \\
\bottomrule
\end{tabular}

% src: vllm/paper_fill.py -> results/b300_tables.json serving; AR ladder f_ld_base tok/s c1=329, c4=1093, c8=1854, c16=2853, c64=5638, c128=8884. Prompt pool: logs/paper/b300/data/4b/pool.jsonl (differs from tab:main).
% The RTX PRO table (4B/9B/27B chain cutoffs + 4B tree) is replaced; B300 ladders exist for 4B only.

\end{table}

\paragraph{Vocabulary shortlisting.}
On the eight-domain 4B benchmark ($8\times4$ tree, batch one), restricting the candidate head to the
shortlist leaves throughput and acceptance nearly unchanged (540.4 vs 541.3 tokens/s, $\tau$ 2.778);
two-pass rescoring of the top $M=\TopM$ candidates raises throughput to 611.7 tokens/s
(13\%) at $\tau$ 2.775. Reducing the budget to $M=1024$ adds
1.1\% throughput at $\tau$ 2.764. Conversely, using $M=\TopM$
instead of 1024 increases acceptance length by 0.011 tokens per round
(0.4\%).
% src: b300_tables.json shortlist (f_sl_full / f_sl_short x2, top-8192 = tab:main tree_1 x3,
% f_sl_m1024 x2; one NVIDIA B300).

% ---------------------------------------------------------------------
% ---------------------------------------------------------------------
\section{Conclusion}
\label{sec:conclusion}

This study presented \Sys, a speculative drafter built on continuous flow-matching
over the target's hidden states. A single Euler step yields the highest
served throughput, since further steps add little acceptance for their
drafting time. Because a variational autoencoder sets flow cost by its
latent width, flow integration takes 0.30--0.38\,ms from 4B to 27B while the
autoregressive step grows from 3.1 to 11.2\,ms, and the pooled speedup peaks at 2.12$\times$ at 27B.
With tree drafting, \Sys outperforms EAGLE-3 and DFlash by 39\% and 11\% at
matched drafter size and training data, and ablations on tree drafting show
that a compact tree can achieve higher speedups than larger, deeper ones.
Adapted in sequence to Turkish, German and French, \Sys leads both baselines
in every language, and the gap widens as new languages are added.

The broader contribution of this work is a formulation of speculative
drafting as transport over the target's own representations, together with
evidence that this formulation holds up across model scales and languages.
The formulation exposes design axes specific to flow-based drafting: where
the transport starts, the latent space it operates in, and the feedback that
restores token dependencies. Load-aware verification and other model
families are natural next steps.

\section*{Limitations}

Like other speculative decoding methods, \Sys gains less at higher concurrency, since larger batches leave less idle compute for verifying drafts. Each serving configuration uses a fixed draft length and tree topology, and while the decode-batch cutoff disables speculation under load, it does not adjust how many tokens are proposed. Budget-aware methods such as BASTION~\citep{oh2026bastion} and EAGLE-4~\citep{eagle4team2026} adapt draft trees to confidence or execution cost, and adding such adaptation to \Sys is future work. The VAE adds a second trade-off. A compact latent space keeps the flow cheap to train and run, but may lose information needed to reconstruct the target’s hidden states, which caps acceptance. A larger latent could help at the cost of slower drafting, a trade-off we have not characterized.

More thorough evaluation would strengthen our conclusions. All experiments use Qwen3.5 targets, whose hybrid attention required engine-specific state handling, so how well \Sys transfers to other model families remains open. Several ablations were run only on a 0.8B target for cost. Our drafters may be undertrained: training on more data would show how the different approaches scale, and pairing larger targets with larger drafters is another open direction.

Tree verification requires a patched engine and greedy decoding. In the
B300 runs, 4--13\% of prompts produce different outputs from plain greedy
decoding across target sizes and languages, with differences attributed
to fp16 near-ties during batched verification. EAGLE-3 and DFlash also
produce differing outputs on the same evaluation prompts. These served
measurements therefore do not establish bitwise equivalence to plain
decoding.
% EN differences /200: 4B 16, 9B 18, 27B 13; TR: 4B 26, 9B 15, 27B 7--8.
% Baselines EAGLE-3/DFlash at 4B: EN 13/13, TR 31/31; plain vs itself 0/200.

\bibliographystyle{mlsys2025}
\bibliography{custom}

\clearpage
\onecolumn
\appendix

\section{Training Data}
\label{sec:appendix-data}

ChainFlow is an efficient and adaptable speculative decoding method, and its adaptability is most visible in the speedups it delivers across domains and languages. To demonstrate this plasticity, we train on four languages: English, Turkish, German and French. We draw on multiple datasets per language, which together cover a wide spectrum of domains: mathematics, code, chat, multi-turn dialogue, retrieval-augmented generation, summarisation, translation and creative writing. Table~\ref{tab:data} lists every source.

\begin{table}[htbp]
\centering\small
\setlength{\tabcolsep}{3pt}
\caption{Training sources by language and domain. Drafters are trained on English first, then adapted in sequence to Turkish, German and French without English replay. English totals 45,887 rows (16.6M tokens), while Turkish, German and French 8,000 each. Replay stages add 2,088 Turkish rows (20.7\%). Rows are prompts, and training responses are the target's own greedy continuations.}
\label{tab:data}
\vskip 0.1in
\begin{tabular}{lllr}
\toprule
Language & Source & Domain & Rows \\
\midrule
English & Code Alpaca~\citep{chaudhary2023codealpaca} & Code & 5,000 \\
        & Dolly~\citep{conover2023dolly} & Instruction & 5,000 \\
        & Nemotron v1, math~\citep{nathawani2025nemotron} & Math & 5,000 \\
        & Nemotron v1, STEM~\citep{nathawani2025nemotron} & Science & 5,000 \\
        & GSM8K~\citep{cobbe2021gsm8k} & Math & 5,000 \\
        & UltraChat~\citep{ding2023ultrachat} & Dialogue & 5,000 \\
        & No Robots~\citep{norobots2023} & Instruction & 4,000 \\
        & WritingPrompts~\citep{fan2018writingprompts} & Creative writing & 3,960 \\
        & CNN/DailyMail~\citep{nallapati2016cnndm} & Summarisation & 4,000 \\
        & OPUS Books, EN$\rightarrow$FR~\citep{tiedemann2012opus} & Translation & 3,927 \\
\midrule
Turkish & InstrucTurca~\citep{instructurca2024} & Instruction & 5,278 \\
        & Function calling~\citep{atasoglu2024funccall} & Function calling & 1,600 \\
        & Tool calling~\citep{atasoglu2024toolcall} & Tool calling & 481 \\
        & Multi-turn dialogue~\citep{magibu2024multiturn} & Dialogue & 641 \\
\midrule
German  & Alpaca-GPT4-Deutsch~\citep{freedomintelligence2023alpacade} & Instruction & 4,794 \\
        & GPT-4 Self-Instruct German~\citep{causallmselfinstructgerman} & Instruction & 1,692 \\
        & GermanRAG~\citep{aurasgermanrag} & RAG & 846 \\
        & OASST2 (de)~\citep{openassistant2023oasst2} & Dialogue & 668 \\
\midrule
French  & French-Alpaca-Instruct & Instruction & 4,956 \\
        & French Instruct ShareGPT & Dialogue & 2,623 \\
        & OASST2 (fr)~\citep{openassistant2023oasst2} & Dialogue & 421 \\
\bottomrule
\end{tabular}
\end{table}

\end{document}